\documentclass[10pt,twocolumn,letterpaper]{article}

\usepackage{cvpr}
\usepackage{times}
\usepackage{epsfig}
\usepackage{graphicx}
\usepackage{amsmath}
\usepackage{amssymb}

\usepackage[pagebackref=true,breaklinks=true,letterpaper=true,colorlinks,bookmarks=false]{hyperref}

 \cvprfinalcopy 

\def\cvprPaperID{****} 

\ifcvprfinal\fi
\begin{document}

\title{Natural Language Person Search Using Deep Reinforcement Learning}

\author{Ankit Shah\\
Language Technologies Institute\\
Carnegie Mellon University\\
{\tt\small aps1@andrew.cmu.edu}
\and
Tyler Vuong\\
Electrical and Computer Engineering\\
Carnegie Mellon University\\
{\tt\small tvuong@andrew.cmu.edu}
}

\maketitle

\begin{abstract}
Recent success in deep reinforcement learning is having an agent learn how to play Go and beat the world champion without any prior knowledge of the game.  In that task, the agent has to make a decision on what action to take based on the positions of the pieces.  Person Search is recently explored using natural language based text description of images for video surveillance applications \cite{2017arXiv170205729L}. We see \cite{2017arXiv170307579W} provides an end to end approach for object based retrieval using deep reinforcement learning without constraints placed on which objects are being detected. However, we believe for real world applications such as person search defining specific constraints which identify person as opposed to starting with a general object detection will have benefits in terms of performance and computational resources requirement. \\ 
In our task, Deep reinforcement learning would localize the person in an image by reshaping the sizes of the bounding boxes. Deep Reinforcement learning with appropriate constraints would look only for the relevant person in the image as opposed to an unconstrained approach where each individual objects in the image are ranked. For person search, the agent is trying to form a tight bounding box around the person in the image who matches the description.  The bounding box is initialized to the full image and at each time step, the agent makes a decision on how to change the current bounding box so that it has a tighter bound around the person based on the description of the person and the pixel values of the current bounding box.  After the agent takes an action, it will be given a reward based on the Intersection over Union (IoU) of the current bounding box and the ground truth box.  Once the agent believes that the bounding box is covering the person, it will indicated that the person is found.

\end{abstract}

\section{Introduction}

Current state of the art techniques for Person search \cite{2017arXiv170205729L} assumes the availability of a dataset where an image or a set of images are cropped to the same size as the candidate person they are trying to find. Further, state of the art techniques for object detection analyzes large amount of region proposals.  Other relevant methods such as object detection using an end-to-end approach with deep reinforcement learning \cite{2017arXiv170307579W} is able to detect persons, however authors believe that their work can be improved upon using constraints for specific tasks such as person identification in our case. \cite{8019485} discusses the benefits of the large scale person re-identification and explores a solution based on Convolutional neural networks to generate a compact descriptor for a coarse search achieving high performance on two existing datasets. We observe that \cite{2016arXiv160401850X} explores a deep learning framework to jointly detect pedestrian detection and person re-identification with a new online instance matching loss function to validate the results. These papers are published regularly with little work done using Deep Reinforcement Learning which we plan to explore through our work. 

Our current task does person search using natural language where we evaluate our results on an image from case to case basis given a description of person we are trying to find. We will find an optimal policy for changing the sizes of the bounding boxes to quickly identify the person in an image. Further, an analysis of the system will be done based on how quickly the system is able to recognize the person of interest in the image provided. Mean average precision over all the samples in our dataset will be used to determine the performance of our approach. We  will further measure the average number of actions needed to successfully find the person along with the average Intersection over Union (metric to find percentage overlap).  \cite{2016arXiv160401850X}. 

\section{Task Description}

One application of deep reinforcement learning is having an agent learn how to play Atari games without any prior knowledge of the game.  For games, the state is the current frame and based on the raw or preprocessed pixel values, the agent has to make a decision on what action to take in order to maximize some score.  In our task, we are given an image and the natural language description query of the person the agent is trying to find in the image.  The agent is initialized to a bounding box around the whole image and at each step, the agent could either shrink the box, expand the box, move in any of the four directions, or terminate.  After the agent takes an action, we will generate some reward mechanism based on how the new bounding box and the old bounding box compares with the ground truth bounding box.  After each action, the agent will update the network.  

We will use Deep Reinforcement Learning to help the agent find the bounding box around the person of interest.  In our case, we use the natural language description for the person rather than a photo of the person.  The problems that we came up with requires us to figure out how to treat the description query and map it to a form we can work with, how to evaluate the candidate bounding box, what kind of deep reinforcement learning methods should we use, the kinds of inputs to the agent and the reward mechanism we are using.

To handle the natural language description, we used a pre-existing sentence embedding model Paragraph2Vec and to get the image features of the bounded box, we used Alexnet which is a model trained on ImageNet.  After we have the sentence features and the image features, we learned a new mapping that maps the features into the same dimensional space.  Reward mechanisms for pong consist of giving a good score for certain actions if the agent wins, and giving a bad score for certain actions if the agent loses.  Some of them give scores for actions based on some weighted average.  For our reward mechanism, if the action took the agent to a new bounding box where the IoU is higher than the previous bounding box, we gave it a positive reward.  If the IoU decreased, we gave a negative reward.  We describe the procedure in detail in our experiments and further sections \\ 

\section{Our Framework}
\begin{figure}[h]
\includegraphics[width=1.0\linewidth]{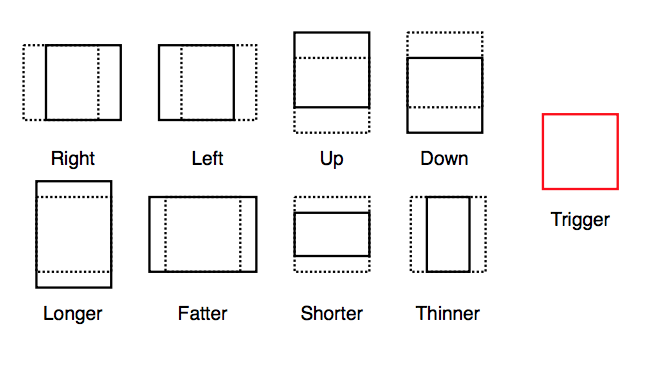}
\caption{Set of all possible Actions}
\label{Action}
\end{figure}

We formulate the task of person search as a Markov Decision Process (MDP). A MDP describes how an agent can be in a set of states, take certain actions, receive rewards based on actions at a given state, and a discount factor used in calculating the cumulative discounted reward which is all donated as \((S,A,R,\gamma )\). where S is the state space, A is the action space, R is rewards for each state. The agent interacts with the environment and at each time step t, the agent takes action \(a_{t}\) based on current state \(s_{t}\) and transitions to the new state \(s_{t+1}\) and receives reward \(r_{t}\) from the environment. It needs the policy function \(\pi(s)\) that specifies what action to take at a given state. One way is to find the optimal action-value function \(Q^{*}(s,a)\), so that the policy  the agent follows is a greedy policy which is at each state, simply choose the action that has the maximum state-action value.  Applying methods such as value iteration to converge to the optimal action-value function is impractical due to a large state space, therefore we use a function approximator to approximate the state-action values. Neural Networks are universal function approximators so the idea is to have a neural network learn the optimal state-action values which we can then use as the optimal policy at each state.  \\
The goal for our agent is to land a tight bounding box around the target person that matches the description. In our case, the state s is the description of the person it is trying to find and the region inside the bounding box at time t in which the agent can transform by taking an action A. We define the following 9 possible actions :  shrink width, shrink height, expand width, expand height, move up, move down, move left, move right and terminate with illustration as seen in Figure \ref{Action}. These are all possible actions which are permissible for our agent and the defined actions are enough to explore the complete environment. \\

The action terminate indicates that the agent believes it has found the person that matches the description.
Reward R depends on the IoU at $S_{t+1}$ compared with $S_{t}$ where S is state of network. IoU is defined as Intersection over Union which is the intersection of the area of the two bounding boxes (predicted and groundtruth bounding boxes) divided by the sum total of the areas of the bounding boxes.  We define b as the bounding box at state s and b' as the bounding box at the new state s' and g as the ground truth box

\begin{equation}
\begin{aligned}
R(s,s') = sign(IoU(b',g) - IoU(b,g))
\end{aligned}
\end{equation}

\begin{equation}
\begin{aligned}
R(terminate) = \begin{Bmatrix} +4 \quad \hspace{8mm} IoU(b,g) \geq .5\\ -2 \hspace{27mm} else \end{Bmatrix}
\end{aligned}
\end{equation}

We formulate the above equation based on following rationale. Our agent receives a positive reward if the agent moves towards the target which in this case is a higher IoU as compared with the previous state. \cite{2016arXiv161103718B} also follows a similar approach and awards a reward of +1 when moving towards the target whereas a reward of -1 when moving away from the target. We define a higher reward for the termination case where the agent thinks it has found the person whereas a lower negative reward is assigned if the agent terminates incorrectly without finding the person or the IoU value is lower than 0.5. 

\section{Deep Q Learning}
A common problem as seen in Reinforcement Learning is that RL is unstable or divergent when an action value function is approximated with a non linear function such as Neural Networks. 
Deep Q learning uses a neural network to approximate the state-action values Q(s,a). Based on the Bellman Equation, the current state-action values should be equal to the immediate reward plus the discounted maximum state-action value at the next state
\[Q(s,a) = R + \gamma max_{a'}Q(s',a')\]
At each time step, the agent looks at the current state S, uses a neural network to estimate the state-action values, takes an action A, receives a reward R and transitions to a new state S'. The network weights are learned by minimizing the mean squared error between the predicted state-action values and the immediate reward plus the discounted maximum state-action value at the next state.  All the information needed to perform that update is represented  in (S,A,R,S')


\begin{figure}[h]
\includegraphics[width=1.0\linewidth]{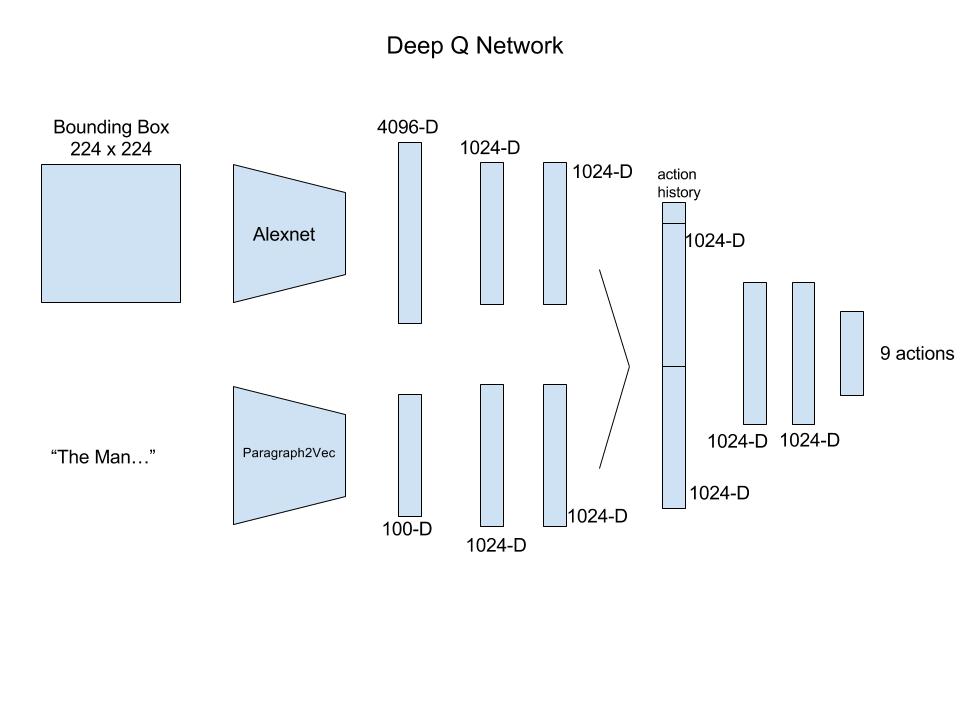}
\caption{Neural Network architecture to predict the next actions}
\label{architecture}
\end{figure}

Rather than optimizing the model based on the current (S,A,R,S'), the idea of experience replay was introduced in order to get rid of correlated updates. Each time the agent chooses an action to transform the box, the current bounding box, action, immediate reward, new bounding box is stored in the experience replay. Then the neural network samples a batch from the experience replay and optimizes the model based on the Bellman Equation. The agent follows an \(\epsilon - greedy\) policy which means that with probability \(\epsilon\), the agent takes a random action and with probability 1 - \(\epsilon\), it takes the action with the highest state-action value. The value of epsilon decreases over time.  Epsilon greedy was used because of the exploration vs exploitation trade off.  In the beginning, rather than letting the agent follow its own policy which is taking the action with the highest state-action value, we make the agent take a random action.  We do this because the network's initial estimates will be far from optimal so we let the network first take random actions.  Once the network has trained enough, the agent starts to follow its own policy rather than just taking random actions hence the value of epsilon decreasing over time.  A pre-trained image model Alexnet was used to transform the current bounding box into a 4096-D feature space and a pre-trained sentence model paragraph2vec was used to transform the sentence descriptions into a 100 - D feature space. We concatenate the image representation, sentence representation, along with a 90-D vector that is simply concatenation of 1-hot encodings of the previous 10 actions.  The output of the model should be a 9 dimensional vector representing the state action values for all 9 actions as shown in Figure \ref{architecture}.

\section{Training}
To train our model, we take in one image at a time and use the same image multiple times before we get a new image.  Each time we let the agent "play" an image, we consider that an episode.  We initially let the agent play the same image for multiple episodes before moving to a new image.  For each episode, the bounding box is initialized to the full image and the agent follows the epsilon greedy policy as mentioned above.  As the agent continues to play the same image, the value epsilon decreases which allows the agent to start following its own policy rather than taking random actions.  Each time the agent is at state s, takes action a, moves to state s', and receives reward r, (s,a,r,s') is all pushed into the replay memory.  After the agent takes an action and transitions into a new state, we then randomly sample a batch from the replay memory and optimize our model based on the sampled batch rather than the immediate action.  That allows for uncorrelated and random updates that will prevent the network from becoming unstable.  After the image has gone through a certain amount of episodes, we obtain a new image.  We repeat the same procedure for all images and continue until we are done training.  After a certain amount of epochs, we start to decrease the initial value of epsilon and the amount of episodes for each image.  

\section{Experiments and Results}
\begin{figure}[h]
\centering
\begin{minipage}{.5\linewidth}
    \includegraphics[width=\linewidth]{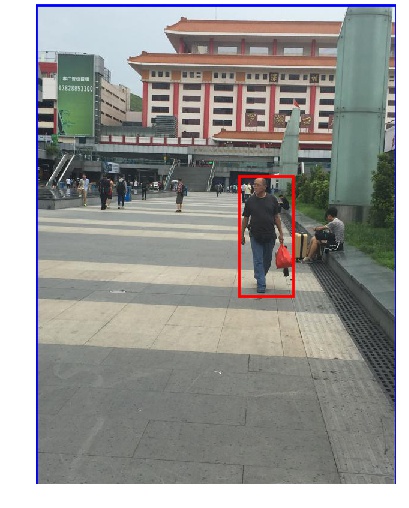}
    \caption{Bounding Box location at t (action) = 0}
    \label{Example 1}
\end{minipage}
\begin{minipage}{.5\linewidth}
    \includegraphics[width=\linewidth]{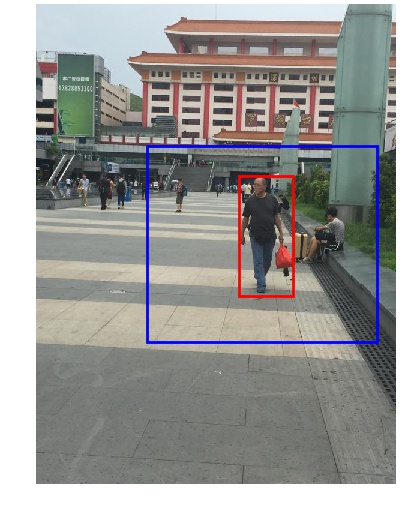}
    \caption{Bounding Box location at t (action) = 7}
    \label{img1}
\end{minipage}
\end{figure}

We perform various experiments to validate our approach. Given the amount of compute power we had at our disposal, we train on a smaller batch of images and perform testing on 100 images to report the results. Training took about 12.5 hours on average for 25 epochs using a NVIDIA TITAN X GPU. Figure \ref{Example 1} 

\begin{figure}[h]
\centering
\begin{minipage}{.5\linewidth}
    \includegraphics[width=\linewidth]{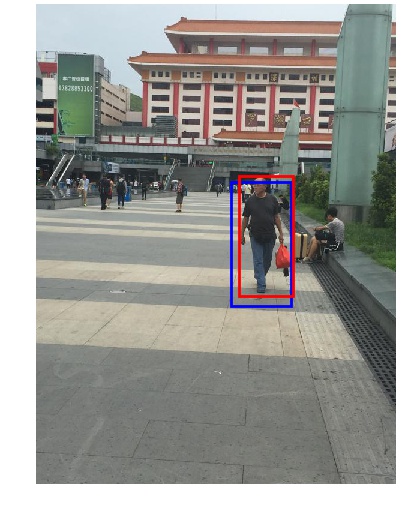}
    \caption{Bounding Box location at t (action) = 15}
    \label{img2}
\end{minipage}
\begin{minipage}{.5\linewidth}
    \includegraphics[width=\linewidth]{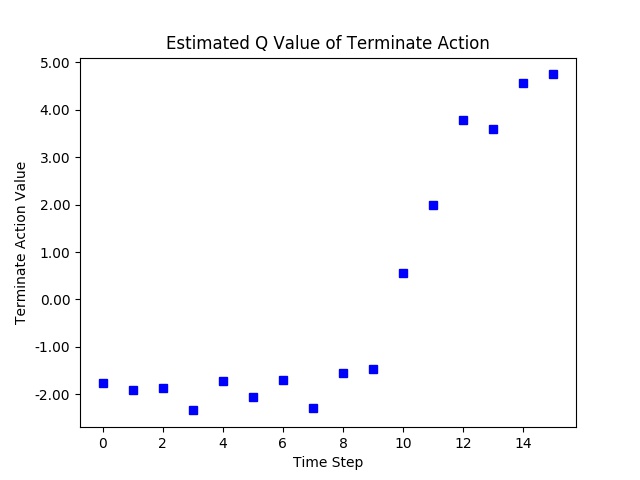}
    \caption{Q(s,9)}
   \label{img3}
\end{minipage}
\end{figure}

\begin{table}[h]
\centering
\caption{Person Search Results}
\label{my-label1}
\resizebox{\columnwidth}{!}{%
\begin{tabular}{|l|l|l|l|l|l|l|}
\hline
                                       & 2 Epochs & 7 Epochs & 10 Epochs & 15 Epochs & 20 Epochs & 25 Epochs \\ \hline
Total Terminated                       & 0        & 0        & .51       & .51       & .69       & .61       \\ \hline
Correctly Terminated                   & 0        & 0        & .67       & .86       & .82       & .95       \\ \hline
Avg IoU                            & .201     & .302     & .387      & .435      & .482      & .485      \\ \hline
Avg IoU Terminate                  & N/A      & N/A      & .528      & .576      & .577      & .591      \\ \hline
Avg IoU No Terminate               & .20      & .302     & .24       & .288      & .271      & .318      \\ \hline
Avg Number Actions 				   & N/A      & N/A      & 15        & 15        & 15.7      & 16.5      \\ \hline
\end{tabular}
}
\end{table}

We observe from Table \ref{my-label1} that the performance increases with correct termination as the number of epochs of training increases. This suggests that we can receive even better performance metrics with a larger number of epochs during the training phase. We receive an average IoU of 0.591 with 25 epochs training which is a reasonable average IoU for the given task. Correctly terminated shows the proportion of the terminated bounding boxes that have an IoU greater than or equal to .5. Average IoU No Terminate shows the IoU of the bounding box at the very last time step without the agent using the terminate action. 

\begin{table}[h]
\centering
\caption{Performance with Different Descriptions}
\label{my-label2}
\resizebox{\columnwidth}{!}{%
\begin{tabular}{|l|l|l|l|}
\hline
                         & Regular Descriptions & Random Descriptions & No Descriptions \\ \hline
Total Terminated         & .61                  & .24                 & .64             \\ \hline
Correctly Terminated     & .95                  & .88                 & .72             \\ \hline
Avg IoU              & .485                 & .358                & .438            \\ \hline
Avg IoU Terminate    & .591                 & .573                & .527            \\ \hline
Avg IoU No Terminate & .318                 & .290                & .278            \\ \hline
Avg Number Action    & 16.5                 & 20                  & 16              \\ \hline
\end{tabular}
}
\end{table}

To further, validate our approach we perform experiments with regular descriptions, random description and no description as sentence vector representation. We observe that our model performs well as seen in Table \ref{my-label2} with random description having over 95 percent correct termination given that the model terminated. The ratio of terminations is huge between providing the regular description vs the random description provided as the input sentence feature vector.

\section{Conclusion}
We demonstrate that a deep reinforcement approach to natural language person search is possible and provides practical results. We observe that having the description helps the agent become more confident on finding the person over time. We obtain approximate 60 percent accuracy for searching the correct person with reasonably small number of actions (16) on average. Our approach is not perfect and even though the agent does not always terminate, it learns to crop out the background and to focus only on people which is a major breakthrough in itself. We plan to extend our work with running more training epochs as well as exploring methods such as Double DQN to form a more robust person search system and related methods. 

\small
\bibliographystyle{ieee}
\bibliography{egbib}



    



\section{Appendix}
\noindent Code available on \href{https:\/\/github.com\/ankitshah009\/11785-Introduction_to_Deep_Learning_Project}{GitHub - Private Repository} \\
Link to our \href{https://drive.google.com/a/andrew.cmu.edu/file/d/17sCN9F3lANV7Y8BwLX84gaJ9n72Ukxe-/view?usp=sharing}{Person Search Demo Video}
\end{document}